\documentclass[conference]{IEEEtran}
\IEEEoverridecommandlockouts
\usepackage{cite}
\usepackage{amsmath,amssymb,amsfonts}
\usepackage{algorithmic}
\usepackage{graphicx}
\usepackage{caption}
\usepackage{subcaption}
\usepackage{textcomp}
\usepackage{url}
\usepackage{array}
\usepackage{multirow}
\usepackage{hyperref}
\hypersetup{
    colorlinks=true,
    linkcolor=blue,
    filecolor=magenta,
    urlcolor=cyan,
    citecolor=blue,
    pdftitle={Calibration and Comparative Analysis of Forward-Looking Sonar and 3D Sonar for Enhanced Underwater Object Recognition},
    pdfpagemode=FullScreen,
    }
\usepackage{xcolor}
\def\BibTeX{{\rm B\kern-.05em{\sc i\kern-.025em b}\kern-.08em
    T\kern-.1667em\lower.7ex\hbox{E}\kern-.125emX}}
\begin{document}

\title{Calibration and Comparative Analysis of Forward-Looking Sonar and 3D Sonar for Enhanced Underwater Object Recognition\\

\thanks{$^{1}$College of Engineering at the University of Florida, Gainesville, FL 32611, USA
{\tt \{apenumarti\}@ufl.edu}}
\thanks{$^{2}$Naval Architecture and Ocean Engineering at the Seoul National University, Seoul 08826, South Korea
{\tt \{janeshin\}@snu.ac.kr}}
\thanks{$^{3}$Electrical and Computer Engineering, University of South Florida, Tampa, FL 33620, USA}
\thanks{$^{4}$University of Miami, Coral Gables, FL 33146, USA}

}

\author{\IEEEauthorblockN{Aditya Penumarti\textsuperscript{1}, Khanh Dong\textsuperscript{1}, Zi-Hao Zhang\textsuperscript{1}, Yongkyoon Park\textsuperscript{2}, Zhenqi Wu\textsuperscript{3}, Trung Dong\textsuperscript{3},\\
Shahriar Negahdaripour\textsuperscript{4}, Xiaomin Lin\textsuperscript{3}, Jane Shin\textsuperscript{2}}
}
\maketitle

\begin{abstract}
Sonars generate a significant amount of noise. With the advent of new technology capable of producing full 3D point clouds, the noise is amplified in sparse point clouds, making it challenging to recognize features for navigation, recognition, or reconstruction. To address this challenge, we propose using two different sonar modalities: one that produces a 2D intensity image and another that generates a 3D point cloud. By implementing auto-calibration, we can filter out noisy features between the modalities to enhance feature extraction. Experiments demonstrate that auto-calibration improves performance over manual calibration by $5\%$ and that filtering enhances feature extraction by more than $40\%$ relative to the raw point cloud. Code and datasets are given at \url{https://theaprilab.org/fls-3d-calibrator}
\end{abstract}


\section{Introduction}
Underwater robotic systems, including autonomous underwater vehicles (AUVs) and remotely operated vehicles (ROVs), have become indispensable in marine operations such as infrastructure inspection, search and rescue, and environmental monitoring. A primary challenge in these missions is the degradation of optical visibility caused by turbidity or the lack of ambient light at depth \cite{mcconnell_perception_2022,sagar_state---art_2025}. Consequently, acoustic sensing often serves as the primary sensing modality for robotic platforms.

Traditional forward-looking sonar (FLS) has been used extensively due to its range and high refresh rate. However, FLS images are obtained by projecting a 3D scene onto a 2D image plane, thereby losing elevation information \cite{gomes_sonar_2025}. This geometric ambiguity often complicates object recognition and spatial reasoning \cite{gomes_sonar_2025,shi_multi-scale_2024,shi_effective_2024}, frequently requiring the vehicle to perform complex maneuvers to capture multiple viewpoints \cite{shin_synthetic_2022}.

Recently, the introduction of 3D echosounders (3D sonar) has shifted the paradigm by providing LiDAR-like point cloud data. Unlike FLS, 3D sonar captures explicit spatial coordinates, offering a direct representation of the environment's geometry \cite{ferreira_underwater_2026,burgul_underwater_2025}. While promising, 3D sonar often faces trade-offs in terms of data density, processing power, and cost compared to FLS. There is currently a gap in the literature regarding the systematic calibration and synergistic use of these two distinct acoustic modalities.

\begin{figure}[t!]
    \centering
    \includegraphics[width=\linewidth]{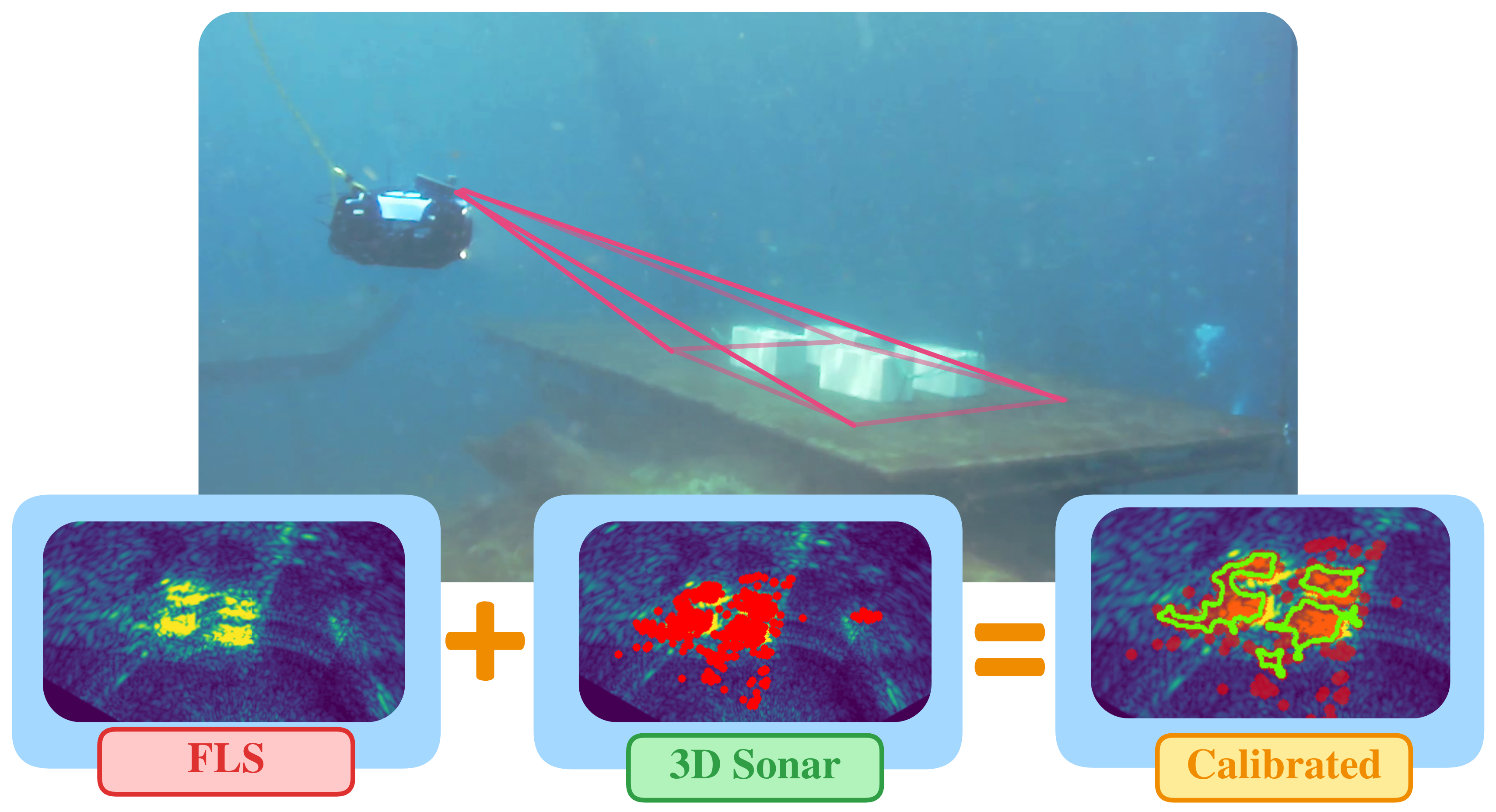}
    \caption{A representative scenario of an ROV scanning a set of 4 cinder blocks. The left frames show the two modalities with their raw data, with the right frame showing the calibrated and segmented objects that were used for auto-calibration.}
    \label{fig:representative_figure}
    \vspace{-15pt}
\end{figure}

In this paper, we propose a novel calibration framework designed to align the coordinate systems of an FLS and a 3D sonar sensor mounted on the same ROV platform. Data were collected during a field test at Blue Grotto in Gainesville, FL, USA, enabling ground-truth estimation from camera images due to the spring's clear water. An overview of the system architecture is given in Fig. \ref{fig:representative_figure}.

\subsection{Contributions}
We introduce three specific contributions in this paper:
\begin{enumerate}
    \item \textbf{Calibration}: An auto-calibration framework that extracts contours on both the FLS and 3D-Sonar and optimizes unknown parameters to find the best calibrated state extrinsics and projection.
    \item \textbf{Denoising}: The calibrated modalities enable the noisy 3D-Sonar point cloud to be denoised for clarity and use in pipelines for reconstruction or recognition.
    \item \textbf{Datasets}: Datasets collected and created are released publicly for reproducibility and training of novel algorithms for similar purposes.
\end{enumerate}
\section{Problem Formulation}\label{sec:problem_formulation}
Consider a multimodal setup with a forward-looking sonar (FLS) and a 3D point-cloud-generating sonar (3D sonar). The FLS produces an image $E(u,v)$ that contains a flattened fan-transformed representation of the sonar array's acoustic distances. The 3D sonar generates a point cloud of form
\begin{equation}
    \mathcal{P} = \{\mathbf{p}_i\}_{i=1}^N , ~\mathbf{p}_i\in \mathbb{R}^3
\end{equation}
where each $\mathbf{p}_i = {[x_i,y_i,z_i]^\top}$ represents a acoustic return in $\mathbb{R}^3$. The two modalities are physically mounted with an arbitrary extrinsic transform between them, and the projection is unknown; the projection is applied to $\mathcal{P}$ to apply filtering and recognition techniques in $I$ owing to superior signal-to-noise characteristics relative to the raw point cloud. Representing the extrinsic and projection parameters via an unknown parameter vector $\boldsymbol{\theta}$, where
\begin{equation}
    \boldsymbol{\theta} = [\underbrace {d_y,\psi}_{\text{extrinsic}},\underbrace{s_r,s_\alpha}_{\text{projection}}]
\end{equation}
where $d_y$ is a translational component in the $y$-axis, $\psi$ is the yaw angle as defined in the FRD (Front-Right-Down) convention coordinate system, and $s_r,s_\alpha$ are scaling components defined by the radius from the sensing origin and azimuth.

Based on the unknown parameter vector and two representative modalities, an arbitrary loss function can be defined:
\begin{equation}
    \mathcal{L}(\boldsymbol{\theta};\mathcal{P}_k,E_{k})
\end{equation}
where $\mathcal{P}_k, E_{k}$ are the pointcloud and FLS image at the current time $k$. This loss function can then be used to solve an optimization calculated over the unknown parameters as:
\begin{equation}\label{eqn:minimization}
    \boldsymbol{\theta}^* = \arg\min_{\boldsymbol{\theta}}\mathcal{L}(\boldsymbol{\theta};\mathcal{P}_k,E_{k})
\end{equation}
which will automatically calibrate the two sensors to produce an aligned representation for object recognition.

\textit{\textbf{Problem Statement}}: Given an acoustic point cloud $\mathcal{P}_k$ and forward looking intensity image $E_{k}$, using an unknown parameter vector $\boldsymbol{\theta}$ to represent the transformation parameter between the two sensing modalities, minimize the cost function $\mathcal{L}(\boldsymbol{\theta};\mathcal{P}_k, E_{k})$ to find the optimal parameters for a calibrated representation $\boldsymbol{\theta}^*$.
\section{Methodology}
\subsection{Projection}
The two modalities are as listed in Sec. \ref{sec:problem_formulation}, including a 2D image and a 3D point cloud. In modern camera-to-LiDAR processing pipelines, the projection involves transforming camera pixels into the LiDAR frame \cite{liu_bevfusion_2023,philion_lift_2020}. Here, we approach the problem in a traditional manner, in which the point cloud is applied to the image \cite{shin_direct_2018}. This intention was implemented by flattening the 3D acoustic returns in an FLS to emphasize object locations in the sensor frame, rather than simply returning the distance to every acoustic return, as with 3D sonar, which would yield a noisy modality.

Initially, the transformation was calculated with an identity extrinsic and unit scaling. This set the unknown parameters to $\boldsymbol{\theta} = \left[0,0,1,1\right]^\top$. The acoustic points are then transformed into the $I$ image plane:
\begin{equation}
    \begin{bmatrix}
        x_{f,i}\\y_{f,i}
    \end{bmatrix} = \begin{bmatrix}
        \cos\psi & -\sin\psi\\\sin\psi & \cos\psi
    \end{bmatrix}\begin{bmatrix}
        x_{\text{pcl},i}\\y_{\text{pcl},i}
    \end{bmatrix} + \begin{bmatrix}
        0\\d_y
    \end{bmatrix}
\end{equation}
where $x_{f,i},y_{f,i}$ are the transformed coordinates in the image plane and $\{x_{\text{pcl},i},y_{\text{pcl},i}\} \in \mathbf{p}_i$.

As the FLS returns acoustic data in a polar-coordinate fan image, $\mathcal{P}$ is transformed into polar coordinates to maintain compatibility.
\begin{equation}
\begin{split}
    r_i &= \sqrt{x_{f,i}^2+y_{f,i}^2}\\
    \alpha_i &= \arctan(y_f,x_f)
\end{split}
\end{equation}
where $r$ is the range or radius, and $\alpha$ is the angle or azimuth.

The polar coordinate transform is then projected into the image coordinate from the FLS $E$:
\begin{equation}
        x_{c,I}^{[i]} = \frac{1}{2}w, ~
        y_{c,I}^{[i]} = h,~
        r_{\text{px},i} =  \frac{hs_rr_i}{r_{\max,E}}
\end{equation}
where $h,w$ are the height and width of the image $E$, $r_{\text{px}}$ is the radius in pixels, and $r_{\max,E}$ is the maximum radius of acoustic returns set on the FLS.

Then the coordinates are transformed into pixels for the image space:
\begin{equation}
    \begin{bmatrix}
x_{\text{px},i}\\y_{\text{px},i}\end{bmatrix} = \begin{bmatrix}
        x_{c,E}^{[i]}& \sin(\alpha_i s_\alpha)\\y_{c,E}^{[i]}& \cos(\alpha_i s_\alpha)
    \end{bmatrix}\begin{bmatrix}
         1\\r_{\text{px},i}
    \end{bmatrix}
\end{equation}

Now this projected space of pixel coordinates is defined as $I_\mathcal{P} = \left\{x_{\text{px},i},y_{\text{px},i}\right\}_{i=1}^{N}$, which will be used to extract contours for the point cloud calibration.

\subsection{Contour Extraction}
Once we have two images from which contours can be extracted, a simple process is implemented to find a thresholded convex hull. Here, the target object used for calibration is a rectangular cinder block, which guides our algorithm development.

The FLS is converted to grayscale $ E \xrightarrow{\mathcal{G}}E_{\text{gray}}$, then thresholded to preserve high-intensity returns.
\begin{equation}
    \tau = \gamma E_{\max}\\
\end{equation}
where $\tau$ is the threshold, $\gamma$ is the percentile and $E_{\max}$ is the maximum intensity of the image. A binary mask is formed from the threshold:
\begin{equation}
    \mathcal{M}(x_\text{px},y_\text{px}) = \left\{\begin{matrix}
        1,& E_{\text{gray}} (x_{\text{px}},y_{\text{px}}) > \tau\\
        0, &E_{\text{gray}} (x_{\text{px}},y_{\text{px}}) < \tau
    \end{matrix} \right.
\end{equation}

To extract object boundaries from the binary mask $\mathcal{M}$, we apply the standard topological structural analysis algorithm \cite{suzuki_topological_1985}, which powers OpenCV's \texttt{findContours} \cite{opencv_library}. The algorithm operates through a single sequential raster scan. When a transition between background and foreground pixels is encountered as defined by the mask, the scan temporarily pauses to trace the boundary path using 8-connectivity \cite{salembier_antiextensive_1998}. Each traced contour is assigned a unique identifier and the contours for each frame is comprised in the set $\mathcal{C}_{k,E}$.

The contours are checked for a minimum area and stored for valid contours $\mathcal{C}_{k,E}>\rho$, where $\rho$ is a set minimum contour area.

Similarly, the transformed point cloud image $I_\mathcal{P}$ has the contour extraction applied. However, it undergoes preliminary preprocessing. Initially, the image is Gaussian-blurred with kernel $K_{5\times5}$. This allows the points to be continuous contours rather than discrete points. From here, a Canny edge detector operation \cite{canny_computational_1986} is applied before performing a morphological close operation \cite{van_horebeek_approximation_2001}. These operations produce clear edges in $I_\mathcal{P}$ before filling the areas between them to create a continuous mask. Finally, the contour-finding operation detailed above is applied to identify the continuous contours $\mathcal{C}_{k,\mathcal{P}}$ in the point cloud.
\subsection{Loss Functions}\label{sec:loss_functions}
To define the quality of calibration, four different loss functions were introduced, with a weighted sum used for optimization. The four loss functions are based on 1) the Chamfer distance, 2) the mutual information, 3) the intersection-over-union (IoU), and 4) coverage error. The calculations for each are detailed below.

\subsubsection{Chamfer Distance}\label{sec:chamfer_distance}
The chamfer distance is defined as the measure of dissimilarity between two finite sets of points based on nearest-neighbor distances \cite{borgefors_hierarchical_1988}. Take the two point sets from the contours $\mathcal{C}_{k,E},~\mathcal{C}_{k,\mathcal{P}}$, and a distance metric $d$, the chamfer distance is defined as:
\begin{equation}
c_k(\mathcal{C}_{k,E},\mathcal{C}_{k,\mathcal{P}}) = \frac{1}{|\mathcal{C}_{k,E}|}\sum_{a\in\mathcal{C}_{k,E}} \min_{b\in \mathcal{C}_{k,\mathcal{P}}} d(a,b)
\end{equation}
Often, the distance metric can be as simple as Euclidean distance for simple measurements. Still, we use a k-dimensional tree to compute the distance between the two contours efficiently. The distance between each contour is computed and vice versa, ($c(\mathcal{C}_{k,E},\mathcal{C}_{k,\mathcal{P}})$ and $c(\mathcal{C}_{k,\mathcal{P}},\mathcal{C}_{k,E})$), which are then summed to produce the chamfer loss $\mathcal{L}(c_k)$.

\subsubsection{Mutual Information (MI)}\label{sec:mutual_information}
Mutual information is the statistical dependency between the two contours. It effectively provides the amount of information one modality provides about the other, showing stronger statistical dependence between the two projections when the contours are aligned. Computing this requires a joint intensity probability over the FLS and the point cloud, given the parameter set $\boldsymbol{\theta}$.
\begin{equation}
    p_k(i,j\mid\boldsymbol{\theta}) = P(\mathcal{C}_{k,E}=i,\mathcal{C}_{k,\mathcal{P}}(\boldsymbol{\theta}))
\end{equation}
where $p_k$ is the joint distribution. Then the marginal distributions are constructed as:
\begin{equation}
    \begin{split}
        p_{E,k} (i) &= \sum_jp_k(i,j)\\
        p_{\mathcal{P},k} (i) &= \sum_ip_k(i,j\mid\theta)\\
    \end{split}
\end{equation}
where $p_{E,k}$ is the FLS probabilty, and $p_{\mathcal{P},k}$ is the point cloud probability. Then the mutual information is computed:
\begin{equation}
    MI(\boldsymbol{\theta}) = \sum_i\sum_jp(i,j\mid\boldsymbol{\theta})\log\left(\frac{p(i,j\mid\boldsymbol{\theta})}{p_{E,k}(i) p_{\mathcal{P},k}(j\mid\boldsymbol{\theta})}\right)
\end{equation}
As we are minimizing a loss function, the mutual information loss is defined as $\mathcal{L}(MI) = -MI(\boldsymbol{\theta})$

\subsubsection{Intersection-Over-Union (IoU)}\label{sec:iou}
The intersection-over-union metric measures how much one bounding box overlaps another. A perfect overlap would yield a 100\% IoU, and no overlap would yield 0\%. Usually, this is computed for a prediction in an image relative to a ground-truth label. Here we will treat $\mathcal{C}_{k,E}$ as the ground truth, and $\mathcal{C}_{k,\mathcal{P}}$ as the predicted boundary. From here, the IoU is defined as:
\begin{equation}
    IoU_k(\boldsymbol{\theta}) = \frac{|\mathcal{C}_{k,E}~\cap~\mathcal{C}_{k,\mathcal{P}}|}{|\mathcal{C}_{k,E}~\cup~\mathcal{C}_{k,\mathcal{P}}|}
\end{equation}
Then the loss for the IoU is defined as $\mathcal{L}(IOU) = 1-IOU(\boldsymbol{\theta})$

\subsubsection{Coverage Error}\label{sec:coverage}
Coverage error computes the filled binary masks for both contours and evaluates the fraction of one contour that is covered by the other. The coverage formula is:
\begin{equation}
    C_{E\leftarrow\mathcal{P}} = \frac{\sum_{u,v} \mathcal{C}_{k,E}\mathcal{C}_{k,\mathcal{P}}}{\sum_{u,v}\mathcal{C}_{k,E}}
\end{equation}
Intuitively, this looks similar to IOU; however, the coverage metric is directional, whereas the IOU is symmetric. Here, much like the IOU, the coverage is subtracted from 1 to form a loss, giving $\mathcal{L}(C_{E\leftarrow\mathcal{P}})=1-C_{E\leftarrow\mathcal{P}}$.

\subsection{Optimization}
The optimization scheme combines the individual loss functions to estimate the unknown parameters. The loss functions are weighted by their significance to ensure that no single loss dominates the others. The combined loss is defined:
\begin{equation}
    \mathcal{L}(\boldsymbol{\theta}) = \zeta\mathcal{L}(c_k)+\xi\mathcal{L}(MI)+\chi\mathcal{L}(IOU)+\eta\mathcal{L}(C_{E\leftarrow\mathcal{P}})
\end{equation}
where $S(\boldsymbol{\theta})=\zeta,\xi,\chi,\eta$ are weighting factors $w_m$ for $M$ metrics. The weights per metric $w_m \in S$ are calculated by perturbing the unknown parameter set by $U(-S_0,S_0)$, where $S_0$ is initialized randomly.
\begin{equation}
    \boldsymbol{\theta}_j = \boldsymbol{\theta}_0 + u, ~ u\sim U(-S_0,S_0)
\end{equation}
where $j\in J$ is an instance of perturbations.
The parameters are then summed over the perturbations to find a distance in the parameter space.
\begin{equation}
    D_{\boldsymbol{\theta}} = \sqrt{\sum_j \left(\frac{\boldsymbol{\theta}_j-\boldsymbol{\theta}_0}{{S_0}}\right)^2}
\end{equation}

These distances are then correlated with the perturbations seen in the loss functions $\mathcal{L}(\boldsymbol{\theta})$ through a Spearman correlation \cite{spearman_proof_1904}:
\begin{equation}
    \rho_m = \rho_{\text{Spearman}}(D_{\boldsymbol{\theta}},\mathcal{L}(\boldsymbol{\theta}))
\end{equation}

Based on this correlation, weights are assigned via:
\begin{equation}
    w_m = \frac{\max(0,\rho_m)}{\sum_m \max(0,\rho_m)}
\end{equation}
which are then assigned to $S(\boldsymbol{\theta})$ appropriately.

The optimization routine is based on the loss function and its weights, using Powell's method, a gradient-free optimizer \cite{powell_efficient_1964}. The method iteratively searches for a local minimum in the parameter space by performing a sequence of 1-D linear searches along a set of directions. The search directions are then updated through observing reductions in the loss functions. This process is computed iteratively until convergence or the maximum number of iterations is reached. As Powell is sensitive to initialization and noise, we solve per frame and take the median of the solutions, which rejects outlier frames from poor contours or spurious returns.
\subsection{Denoising Scheme}\label{sec:denoising_scheme}
The denoising scheme is a simple binary accept-or-reject scheme based on contour envelopes. Once calibrated, the projected point cloud still contains points that can be filtered based on the FLS contour. Given the binary Mask expressed earlier $\mathcal{M}$ and using the calibrated projection $I_{\mathcal{P}}$, each point $\mathbf{p}_i \in \mathcal{P}$ is projected into pixel coordinates $(u_i,v_i)$. The set of retained point indices of the mask correspondence is defined as:
\begin{equation}
    I_{\text{mask}} = \{i\mid \mathcal{M}(u_i,v_i)=1\},~ i= 1,
    \dots N
\end{equation}
Then the filtered point cloud is:
\begin{equation}
    \mathcal{P}_{\text{mask}} = \{\mathbf{p}_i\mid i \in I_{\text{mask}}\}
\end{equation}

This filtered point cloud will be used as a validation for object recognition performance to show the benefits of having both modalities present during sensing.

\section{Experiments}
\begin{figure}[t]
    \centering
    \includegraphics[width=\linewidth]{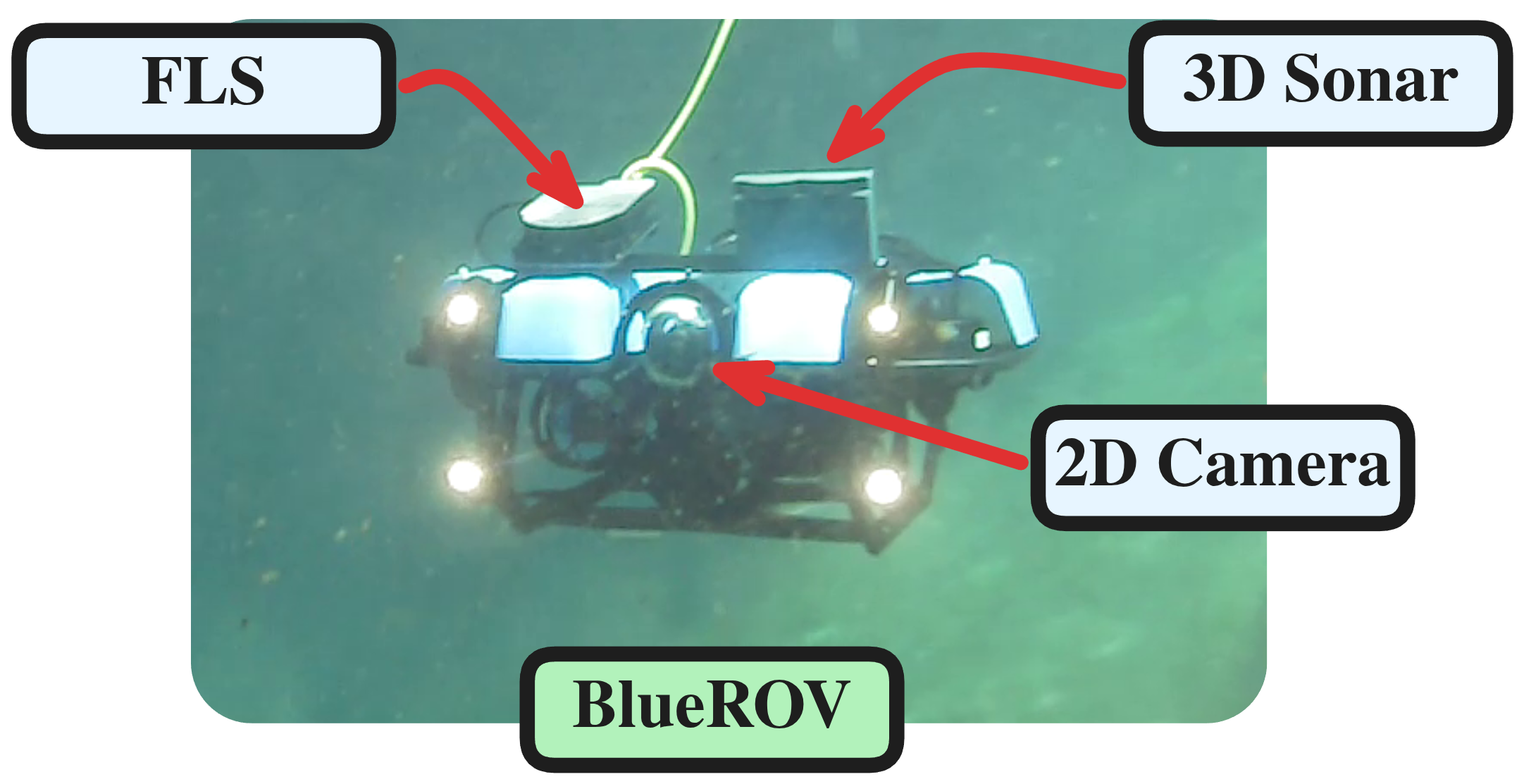}
    \caption{Blue Robotics Blue ROV equipped with a forward-looking sonar and 3D sonar.}
    \label{fig:blue_rov}
    \vspace{-8pt}
\end{figure}
\begin{table}[t!]
\centering
\caption{Comparative results for the automatic calibration versus the manually measured transformation using manually labeled data.}
\label{tab:label_calibration}
\setlength{\tabcolsep}{2pt}
\renewcommand{\arraystretch}{1.0}
\small
\begin{tabular}{>{\centering\arraybackslash}p{1.5cm}>{\centering\arraybackslash}p{2cm}c>{\centering\arraybackslash}p{1.5cm}>{\centering\arraybackslash}p{2cm}}
\hline\hline
Calibration Type
    & \textbf{Dice Score (F1) $\uparrow$}
    & \textbf{IoU} $\uparrow$
    & \textbf{Chamfer} ($\mathrm{px}$) $\downarrow$& \textbf{Reproj. Error} ($\mathrm{px}$) $\downarrow$ \\
\hline
\emph{Manual}
    & $0.761$   & $0.64$ & $4.11$ & $\mathbf{7.17}$ \\
\hline
\emph{Auto}
    & $\mathbf{0.795}$  & $\mathbf{0.68}$ & $\mathbf{3.67}$ & $7.46$ \\
\hline\hline
\end{tabular}
\end{table}

\begin{figure}[t]
    \centering
    \includegraphics[width=0.7\linewidth]{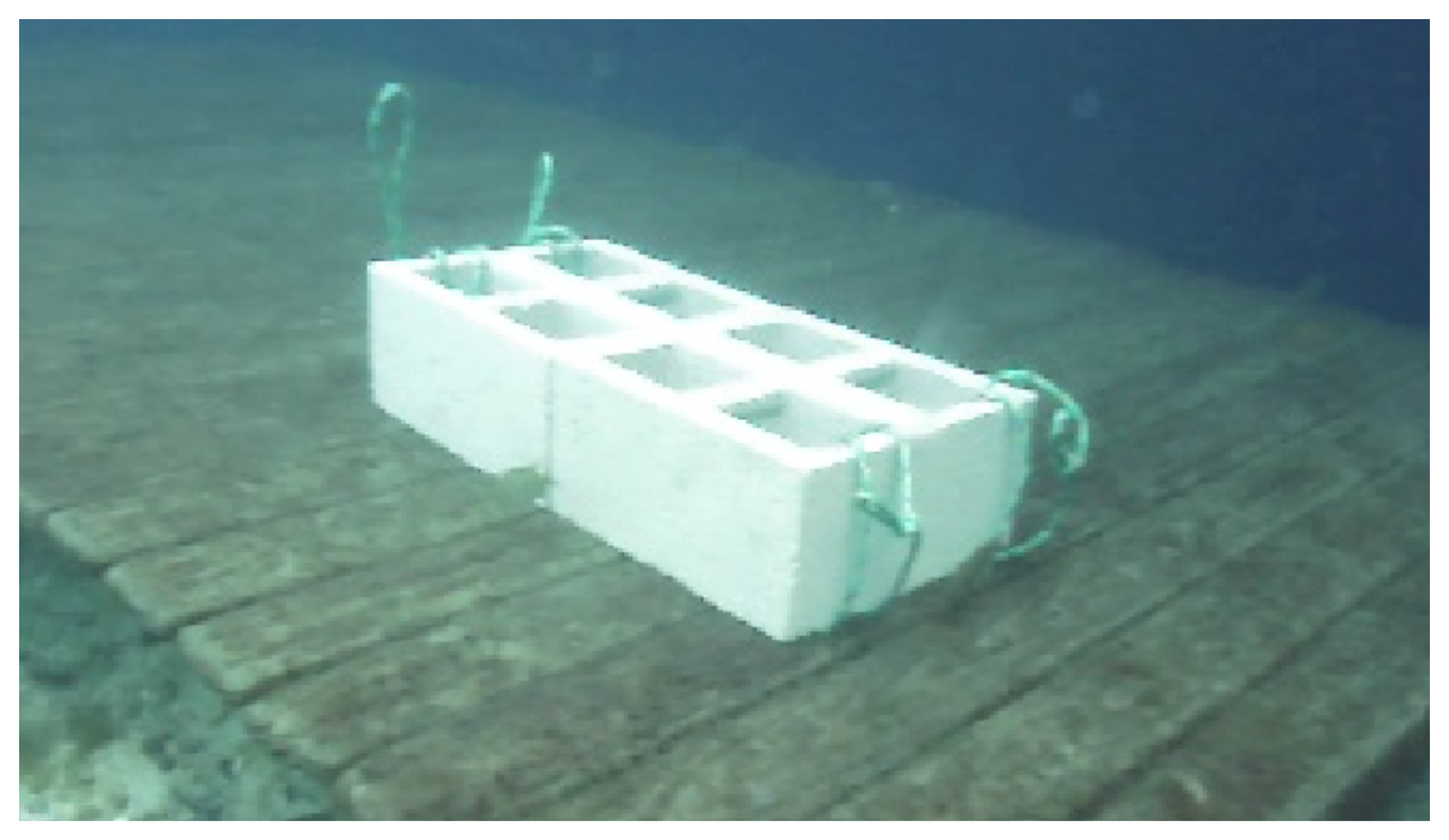}
    \caption{Cinder block placed on a deck underwater at Blue Grotto, Williston, FL, USA. The cinderblock was chosen as the target for calibration due to its bright return on both modalities. Several tests were conducted in different configurations to find the best for calibration.}
    \label{fig:cinderblock}
    \vspace{-8pt}
\end{figure}
\begin{table}[t!]
\centering
\caption{Statistical data for the metrics for the automatic and manual calibration using contour extraction.}
\label{tab:statistical_calibration}
\setlength{\tabcolsep}{2pt}
\renewcommand{\arraystretch}{1.0}
\small
\begin{tabular}{>{\centering\arraybackslash}p{1.5cm}>{\centering\arraybackslash}p{1.5cm}>{\centering\arraybackslash}p{1.0cm}>{\centering\arraybackslash}p{2cm}>{\centering\arraybackslash}p{2cm}}
\hline\hline
Calibration Type
    & \textbf{Coverage} $\uparrow$
    & \textbf{IoU} $\uparrow$
    & \textbf{Mutual Information} $\uparrow$& \textbf{Hausdorff} ($\mathrm{px}$) $\downarrow$\\
\hline
\emph{Manual}
    & $0.59$   & $0.45$ & $0.29$ & $35.39$ \\
\hline
\emph{Auto}
    & $\mathbf{0.64}$  & $\mathbf{0.50}$ & $\mathbf{0.33}$ & $\mathbf{34.89}$ \\
\hline\hline
\end{tabular}
\end{table}
\begin{table}[t!]
\centering
\caption{Object recognition performance on denoised data and cropped raw data. The raw, unfiltered point cloud, the statistical outlier removal (SOR) filter, and the FLS-filtered point clouds are compared.}
\label{tab:pointnet}
\setlength{\tabcolsep}{2pt}
\renewcommand{\arraystretch}{1.0}
\small
\begin{tabular}{>{\centering\arraybackslash}p{2cm}>{\centering\arraybackslash}p{2cm}c}
\hline\hline
Filtering Type
    & \textbf{Dice Score} $\uparrow$
    & \textbf{IoU} $\uparrow$\\
\hline
\emph{No Filter (Raw)}
    & $0.59$   & $0.42$\\
\hline
\emph{SOR}
    & $0.78$  & $0.63$\\
\hline
\emph{FLS-Filtered}
    & $\mathbf{0.85}$  & $\mathbf{0.73}$\\
\hline\hline
\end{tabular}
\end{table}
\begin{figure*}[t!]
    \centering
    \begin{subfigure}{0.28\textwidth}
    \includegraphics[width=\textwidth]{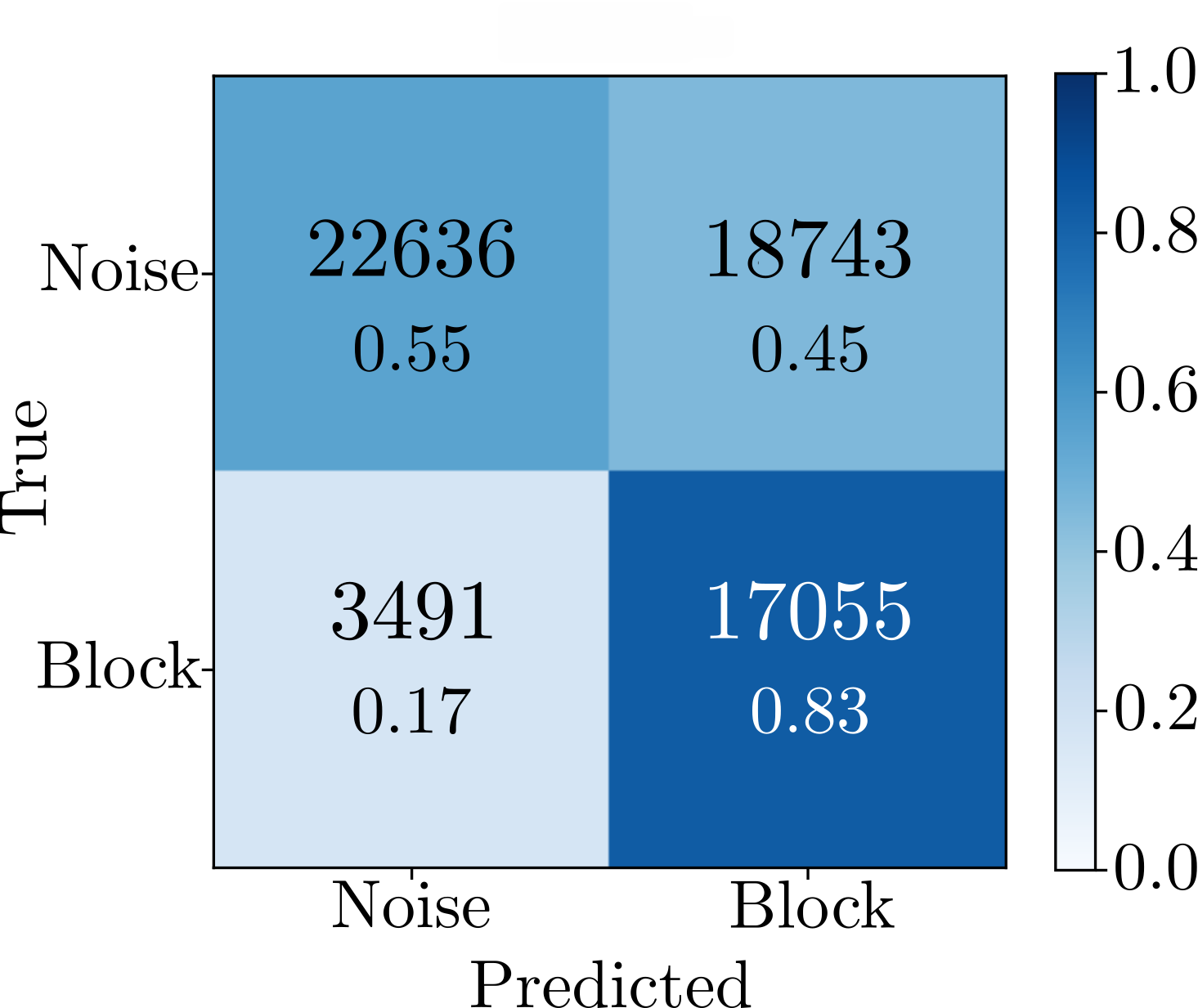}\
    \caption{Raw point cloud}
    \end{subfigure}
    \begin{subfigure}{0.28\textwidth}
    \includegraphics[width=\textwidth]{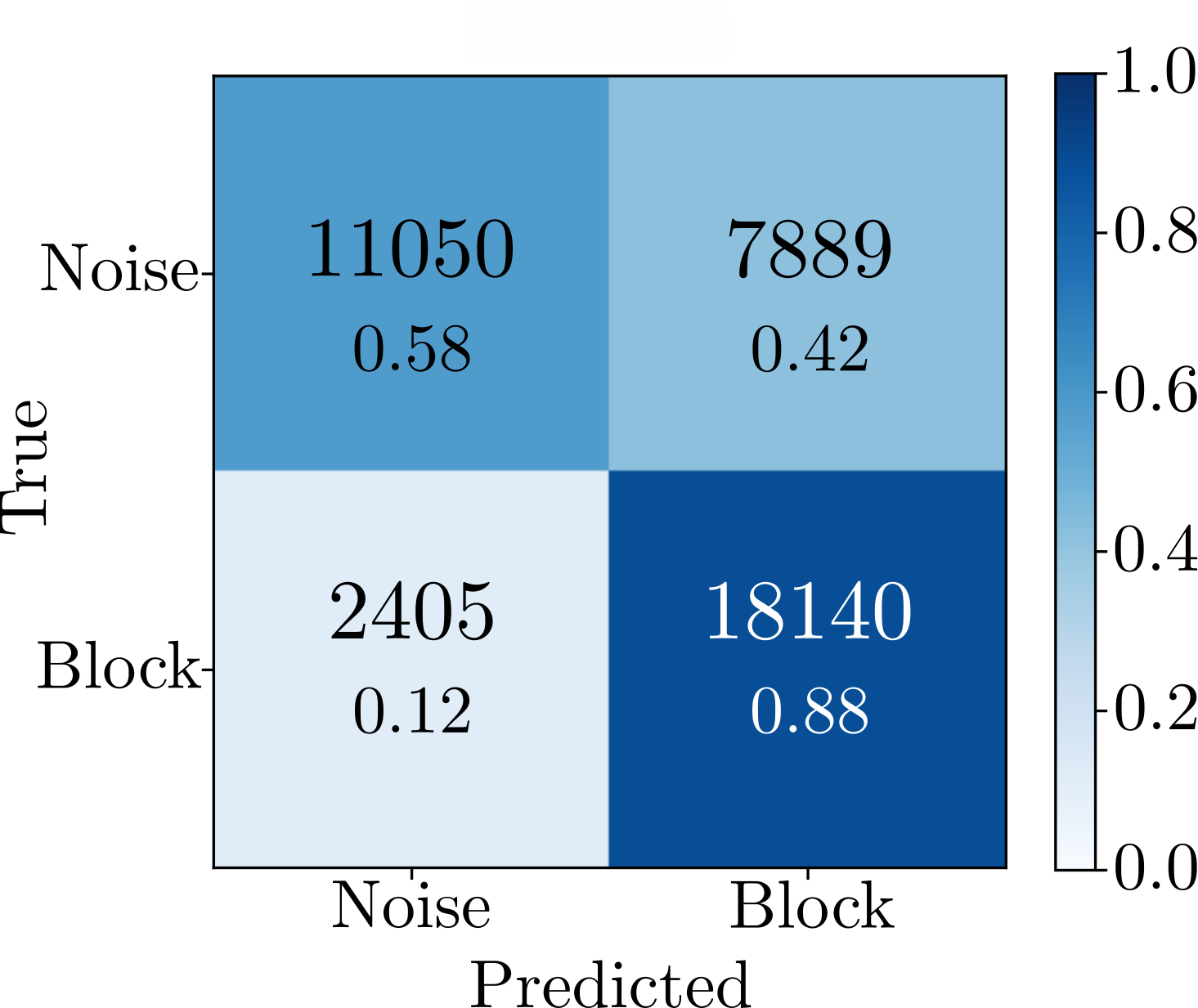}
    \caption{SOR filtered point cloud}
    \end{subfigure}
    \begin{subfigure}{0.28\textwidth}
    \includegraphics[width=\textwidth]{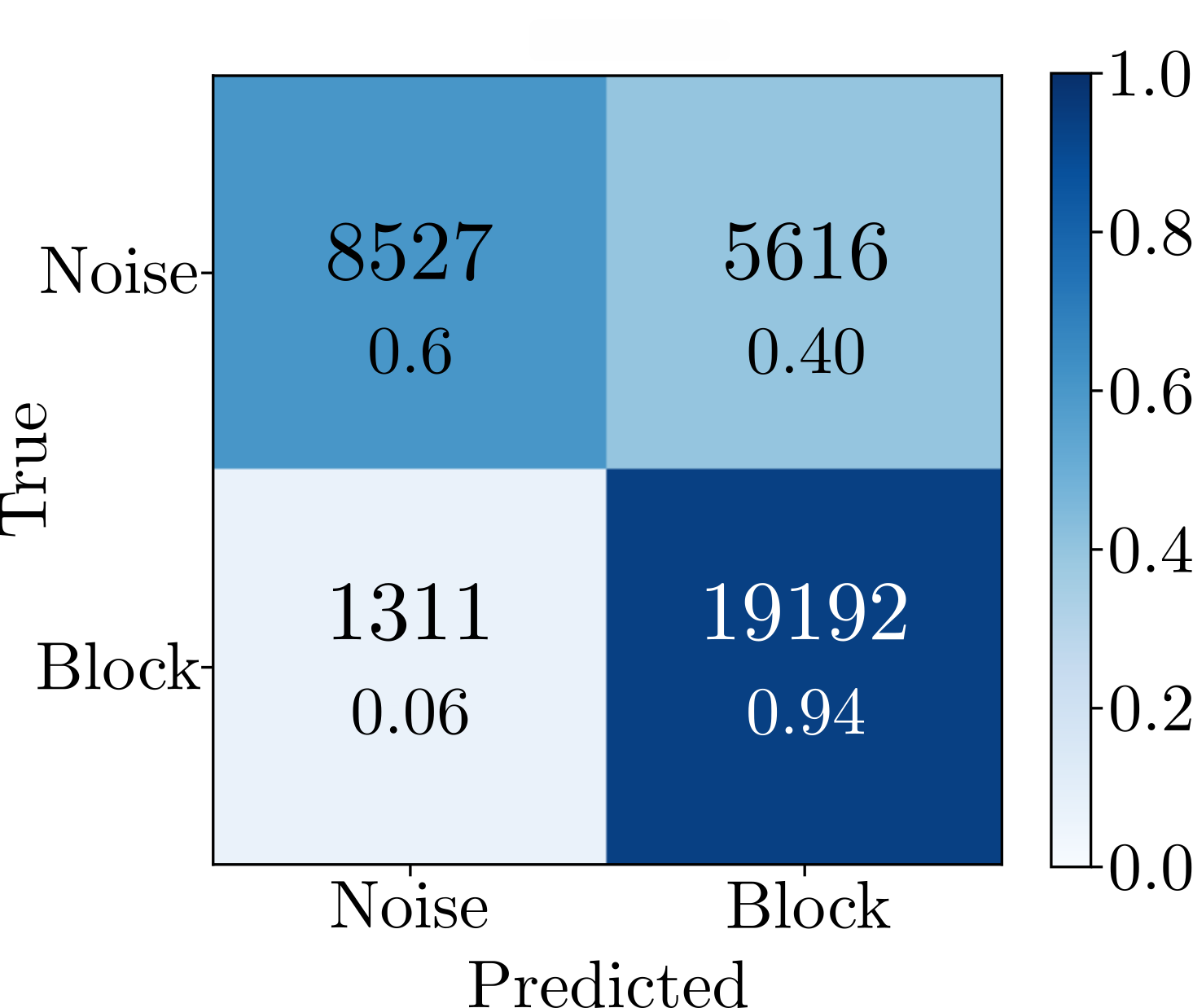}
    \caption{FLS-3D sonar filtered point cloud}
    \end{subfigure}
    \caption{Confusion matrices for the different filtering methods used prior to feature extraction using PointNet. (a) Shows the raw point cloud, which was cropped to the FLS range, (b) is the SOR filtered point cloud, and (c) is the calibrated FLS-3D sonar filtering.}
    \label{fig:confusion_matrix}
    \vspace{-10pt}
\end{figure*}
\begin{figure}[t!]
    \centering
    \begin{subfigure}{0.32\textwidth}
    \includegraphics[width=\textwidth]{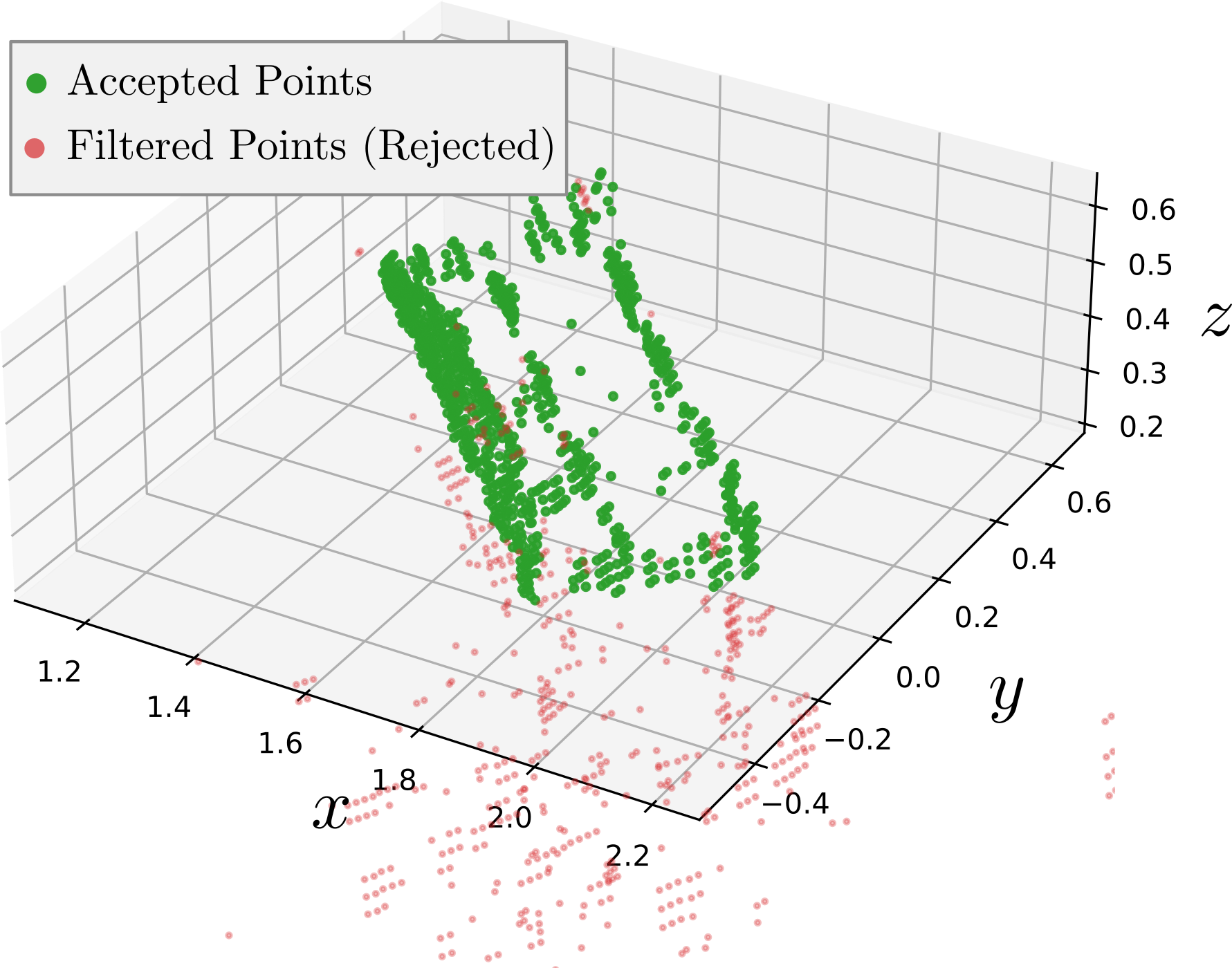}
    \caption{Ground truth label}
    \end{subfigure}
    \begin{subfigure}{0.32\textwidth}
    \includegraphics[width=\textwidth]{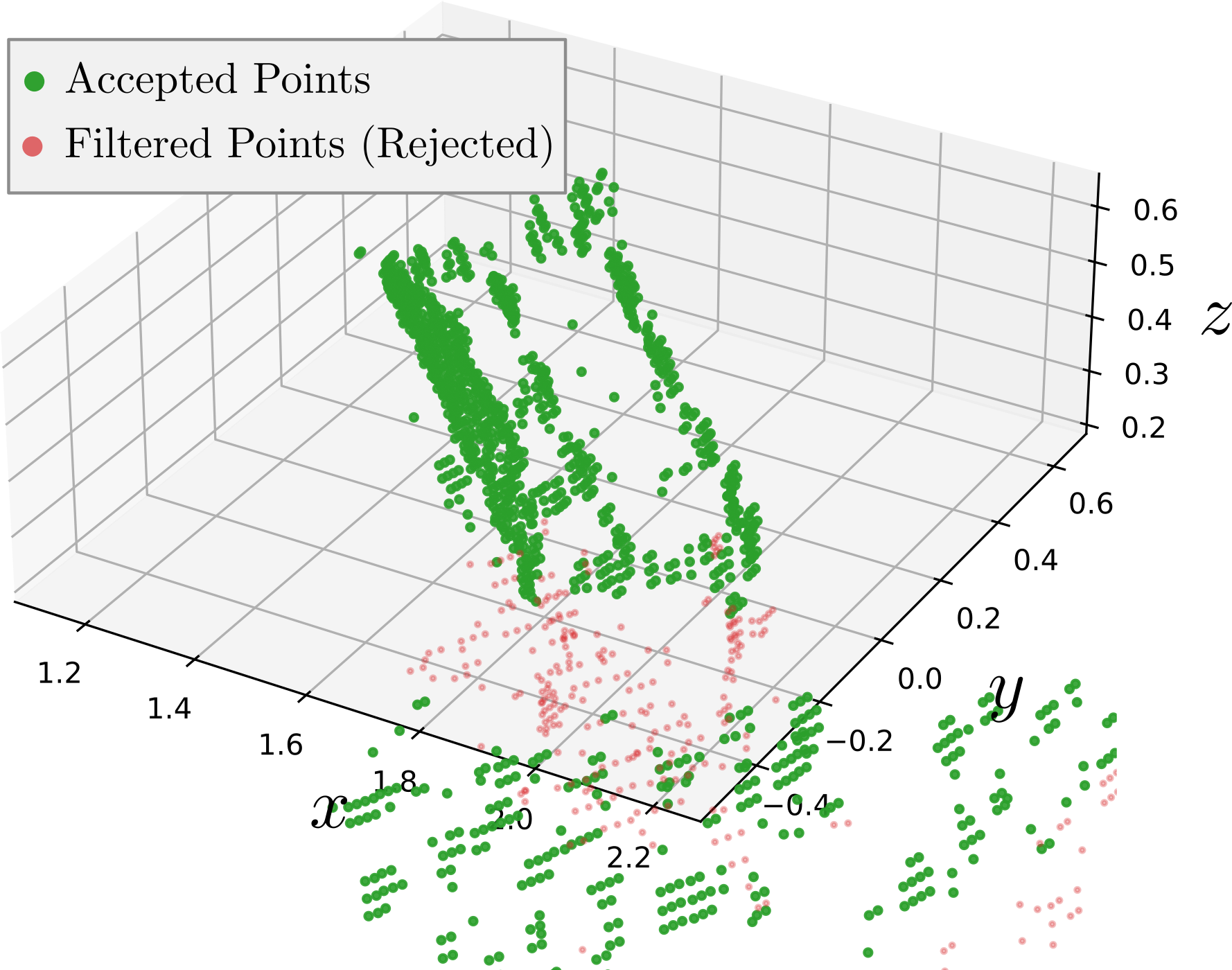}\
    \caption{SOR PointNet segmentation}
    \end{subfigure}
    \begin{subfigure}{0.32\textwidth}
    \includegraphics[width=\textwidth]{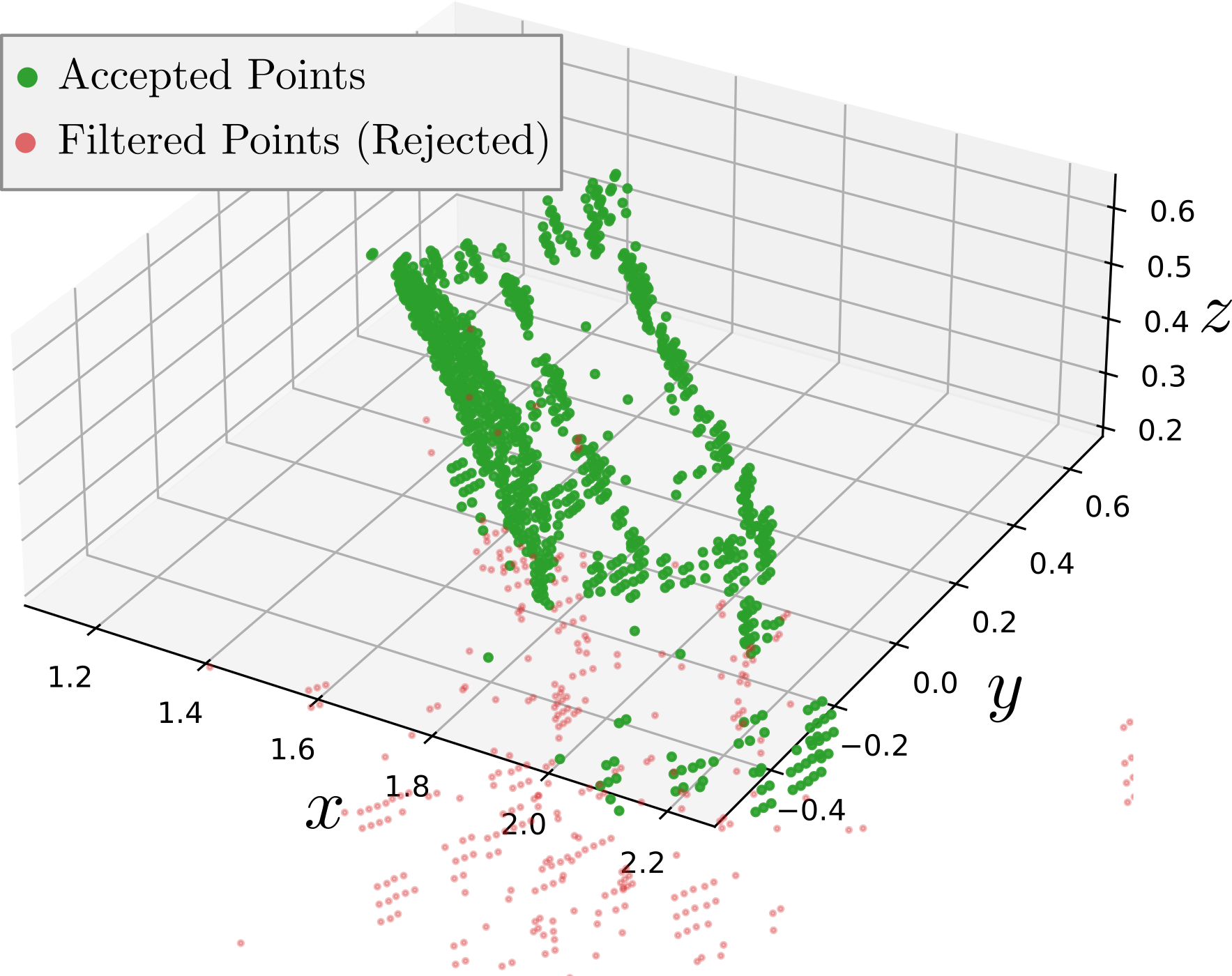}
    \caption{2D-3DSonar PointNet segmentation}
    \end{subfigure}
    \caption{Segmentation results from PointNet. (a) Ground truth label for the PointNet segmentation validation, (b) PointNet results for the SOR filtering, and (c) PointNet results on the FLS-3D sonar filtering.}
    \label{fig:pointnet_segmentation}
    \vspace{-15pt}
\end{figure}
Two experiments were conducted: 1) validation of the calibrated methods against a ground-truth label and a manual calibration comparison, and 2) validation of the denoising process to improve feature recognition.
\subsection{Experimental Setup}
The experiments were conducted using a BlueROV2, an underwater remotely operated vehicle. The vehicle was outfitted with two sonars (FLS and 3D Sonar) and an RGB camera for viewing and tele-operation. A hardware setup is shown in Fig. \ref{fig:blue_rov}. The calibration was conducted on data that was collected from 4 cinder blocks placed underwater (see Fig. \ref{fig:cinderblock}). The data was collected and sorted into $4 ~\mathrm{s}$ snapshots, which were then used for calibration and validation. A small subset of the data was labeled for ground-truth confirmation. The calibration parameters used for conducting experiments are $\boldsymbol{\theta}^* = [-0.2645,-0.0004,1.0294,1.0304]^\top$

The feature extraction experiment was designed to use PointNet \cite{qi_pointnet_2017}, a neural point cloud feature segmentation system, to test capabilities. PointNet was trained on synthetically generated ground-truth data and then tested on data collected by the two sensors. The synthetic clouds model the four cinder blocks in their field arrangement. Each cloud is subsampled, and the faces oriented away from the sensor and toward the substrate are occluded to emulate self-occlusion and acoustic shadowing. Noise is added by drawing surrounding points from a Gaussian distribution centered on the block structure. Candidate points falling within a minimum distance of any block point are rejected, so the noise populates the surrounding volume rather than the object itself. A fraction of the clouds is left undegraded as well, so the network observes the ideal object structure. For model evaluation, both the filtered and raw point clouds are manually annotated using labelCloud \cite{sager_labelcloud_2022} to provide per-point ground truth.
\subsection{Calibration Experiments}
 The statistical validation of calibration quality using labeled data is given in Table \ref{tab:label_calibration}. The metrics shown were the Dice score, which measures how much two sets overlap, IOU, which is described in Sec. \ref{sec:iou}, chamfer (Sec. \ref{sec:chamfer_distance}), and the reprojection error, which describes the projected distance from a pixel to a 3D point.

As shown in Table \ref{tab:label_calibration}, the auto calibration outperforms the manual calibration across all metrics. This is due to the difficulty of measuring the azimuth offset, radius, and azimuth scaling. The auto-calibration inherently allows for the minute changes in scaling and offsets. The calibration performance is approximately a $5\%$ improvement across overlap metrics (IoU and coverage) and a $0.5\mathrm{px}$ improvement in the distance metrics.

Table \ref{tab:statistical_calibration} shows statistical validation over $336$ point cloud frames. The auto-calibration outperforms the manual calibration, but marginally on IOU and MI, even though the IOU and MI are expected to be penalized heavily for edge loss \cite{yang_nonrigid_2019,cheng_boundary_2021}. This occurs mainly due to the equal treatment of edge and area pixels, as well as to discretization. This leads to around a $5\%$ improvement in overlap metrics and a $0.5\mathrm{px}$ distance improvement over the manual calibration.

While the improvements achieved through optimization on our setup are minimal, the ability to fully calibrate the two modalities without a ground-truth measurement still enables efficient on-the-fly calibration.

\subsection{Denoising Object Recognition}
To validate that the calibration-enabled denoising provides tangible benefit beyond geometric alignment, we evaluated its effect on a downstream object recognition task. Specifically, we perform point-wise binary segmentation on the point cloud using a PointNet segmentation network \cite{qi_pointnet_2017}.

We compare three inputs, as reported in Table~\ref{tab:pointnet}. The raw cloud is cropped to the FLS working range of 6 meters. The FLS-filtered condition applies the calibrated denoising scheme shown in Sec. \ref{sec:denoising_scheme}. We apply statistical outlier removal (SOR) \cite{rusu_towards_2008}, a geometric denoiser that rejects points far from their neighbors for additional validation.

Table~\ref{tab:pointnet} shows that FLS-filtering achieves a relative improvement of $40\%$ in Dice and $70\%$ in IoU over the raw cloud and exceeds the geometric SOR result by $9\%$ in Dice and $16\%$ in IoU. Noise lying close to the object is geometrically indistinguishable from the labels and survives SOR, but is rejected by the FLS contour because it lacks a corresponding 2D return. As seen in Fig.~\ref{fig:pointnet_segmentation}, PointNet then segments this residual noise as part of the block. The confusion matrices in Fig.~\ref{fig:confusion_matrix} confirm this, showing that FLS-filtering primarily removes the false positives that dominate the raw and SOR predictions.

\section{Conclusions}
We present a novel multimodal fusion auto-calibration between an FLS and a 3D sonar system that improves overlap by $5\%$ and distance errors by $0.5\mathrm{px}$ in a statistical sense. When a denoised point cloud was extracted using a calibrated representation, segmentation performance improved by over $40\%$ compared to a raw point cloud and over $9\%$ compared to a cropped and SOR-filtered point cloud. While the gains in the tested results are marginal due to the relatively clean data acquired, in highly turbid environments, the calibrated denoising will yield a significant advantage over traditional denoising techniques, allowing for better segmentation and object recognition.
\subsection{Limitations}
While auto-calibration optimization yields better results than manual calibration and better segmentation, it incurs a significant loss of information during calibration by collapsing elevation to facilitate feature matching. The loss of information can lead to poorer calibration filters, either because the 3D objects being scanned cannot be sufficiently reconstructed or because erroneous noise is introduced into the denoising scheme. As future work, it would be preferable to test the system by projecting the FLS image onto the 3D sonar point cloud to preserve elevation information.
\section*{Acknowledgment}
The authors would like to thank Andres Pulido, Nikhil Iyer, Cayman Christ, and Cheryl Thacker for their help with diving, and Blue Grotto Dive Resort for the facilities for data collection. This work was supported by CRADA22-0034-J003,  NVIDIA Academic Grant Program, and the New Faculty Startup Fund from  Seoul National University.

\bibliographystyle{IEEEtran.bst}
\bibliography{references}
\end{document}